\documentclass{article}

\usepackage[preprint]{neurips_2024}

\usepackage[utf8]{inputenc} 
\usepackage[T1]{fontenc}    
\usepackage{url}            
\usepackage{booktabs}       
\usepackage{amsfonts}       
\usepackage{nicefrac}       
\usepackage{microtype}      
\usepackage{xcolor}         
\usepackage{wrapfig}
\usepackage{graphicx}
\usepackage{amsmath}
\usepackage{multirow}  
\usepackage{makecell}
\usepackage{arydshln}

\definecolor{cvprblue}{rgb}{0.21,0.49,0.74}
\usepackage[pagebackref,breaklinks,colorlinks,citecolor=cvprblue]{hyperref}

\newcommand{\ie}{\textit{i}.\textit{e}.}
\newcommand{\eg}{\textit{e}.\textit{g}.}
\newcommand{\etc}{etc\@ifnextchar.{}{.\@}}

\newcommand{\BEST}[1]{{\textcolor[rgb]{1,0,0}{#1}}}
\newcommand{\SBEST}[1]{{\textcolor[rgb]{0,0,1}{#1}}}

\title{LoReTrack: Efficient and Accurate Low-Resolution Transformer Tracking}

\author{%
  Shaohua Dong$^{1}$, Yunhe Feng$^{1}$, Qing Yang$^{1}$, Yuewei Lin$^{2}$, Heng Fan$^{1}$ \\
  $^1$Department of Computer Science and Engineering, University of North Texas \\ $^2$Brookhaven National Laboratory \\
  \texttt{shaohuadong@my.unt.edu, heng.fan@unt.edu}
}

\begin{document}

\maketitle


\begin{abstract}

    High-performance Transformer trackers have shown excellent results, yet they often bear a heavy computational load. Observing that a smaller input can \emph{immediately} and \emph{conveniently} reduce computations without changing the model, an easy solution is to adopt the low-resolution input for efficient Transformer tracking. Albeit faster, this hurts tracking accuracy much due to information loss in low resolution tracking. In this paper, we aim to mitigate such information loss to boost the performance of the low-resolution Transformer tracking via dual knowledge distillation from a frozen high-resolution (but \emph{not} a larger) Transformer tracker. The core lies in two simple yet effective distillation modules, comprising query-key-value knowledge distillation (QKV-KD) and discrimination knowledge distillation (Disc-KD), across resolutions. The former, from the global view, allows the low-resolution tracker to inherit the features and interactions from the high-resolution tracker, while the later, from the target-aware view, enhances the target-background distinguishing capacity via imitating discriminative regions from its high-resolution counterpart. With the dual knowledge distillation, our \textbf{Lo}w-\textbf{Re}solution Transformer \textbf{Track}er (\emph{\textbf{LoReTrack}}) enjoys not only \emph{high efficiency} owing to reduced computation but also \emph{enhanced accuracy} by distilling knowledge from the high-resolution tracker. In extensive experiments, LoReTrack with a 256$^{2}$ resolution consistently improves baseline with the same resolution, and shows competitive or even better results compared to 384$^{2}$ high-resolution Transformer tracker, while running \textbf{52\%} faster and saving \textbf{56\%} MACs. Moreover, LoReTrack is resolution-scalable. With a 128$^{2}$ resolution, it runs 25 \emph{fps} on a CPU with 64.9\%/46.4\% SUC scores on LaSOT/LaSOT$_{\mathrm{ext}}$, surpassing all other CPU real-time trackers. \href{https://github.com/ShaohuaDong2021/LoReTrack}{Code} will be released. 
    
\end{abstract}

\section{Introduction}

Visual tracking aims to continuously locate the target of interest throughout a video. It is a fundamental computer vision problem, and has received extensive attention in the past decades for its crucial roles in many applications including visual surveillance, robotics, and others. With the introduction of Transformer~\cite{vaswani2017attention,dosovitskiy2020image} in recent years, considerable progress has been achieved in tracking community. Many high-performance Transformer trackers (\eg,~\cite{lin2022swintrack,ye2022joint,cui2022mixformer,chen2023seqtrack,wei2023autoregressive}) have been proposed and exhibited previously unattainable accuracy. Despite this, they usually suffer from heavy computational burden, which largely limits their practical deployment to real applications.

\begin{figure}[!t]
    \centering    \includegraphics[width=0.945\linewidth]{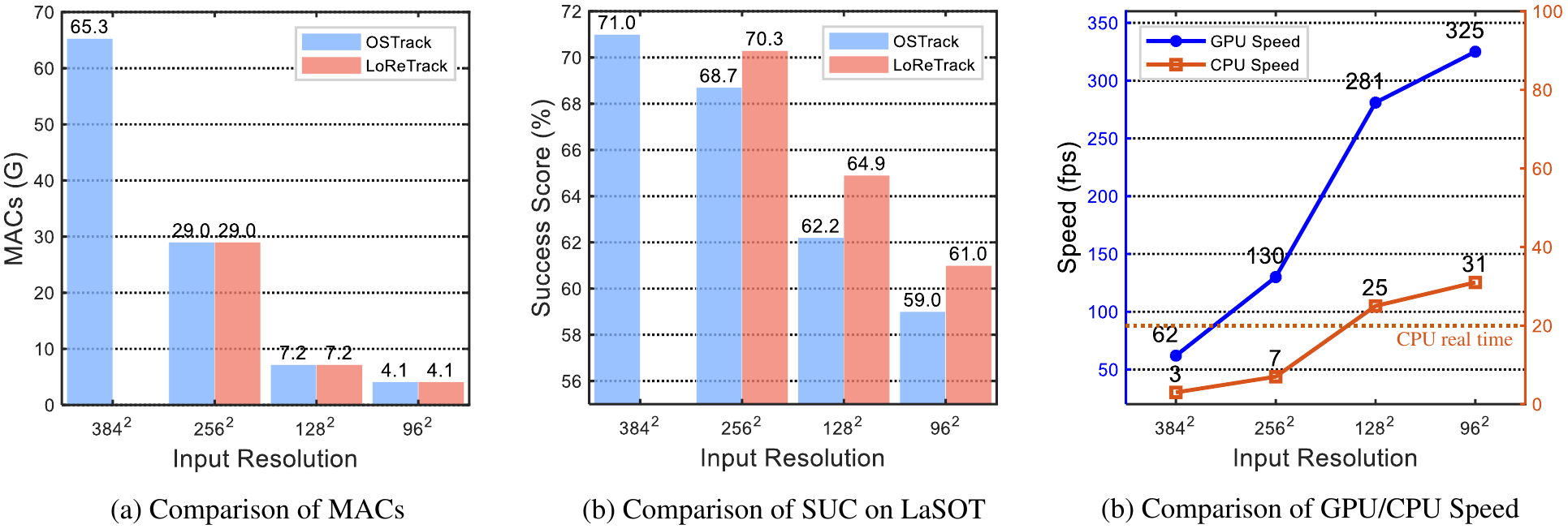}
    \caption{Comparison of computation (a), accuracy (b), and GPU/CPU speed in different resolutions for OSTrack~\cite{ye2022joint} and LoReTrack on LaSOT~\cite{fan2019lasot}. Please note, candidate elimination is not used in both methods. \emph{Best viewed in color and by zooming in for all figures}.}
    \vspace{-10pt}
    \label{fig:1}
\end{figure}

In order to overcome this limitation, existing solutions mostly focus on reducing the model complexity, such as leveraging a lighter Transformer backbone (\eg,~\cite{kang2023exploring}), compressing the Transformer backbone (\eg,~\cite{cui2024mixformerv2}), and developing compact Transformer attentions (\eg,~\cite{blatter2023efficient}), to improve the running efficiency of Transformer trackers. In addition to model complexity, another \emph{crucial factor} affecting computation and tracking efficiency is input resolution~\cite{tan2019efficientnet}. For example, when decreasing input resolution\footnote{For simplicity, input resolution $k^{2}$ (\ie, $k\times k$) indicates the resolution for the search region of a Transformer tracker, and the resolution of its template is usually half that of search region unless otherwise specified.} of the popular one-stream Transformer tracker OSTrack~\cite{ye2022joint}  (\eg, 384$^{2}$$\rightarrow$256$^{2}$$\rightarrow$128$^{2}$$\rightarrow$96$^{2}$), its computation in Multiply–Accumulate Operations (MACs), as displayed in Fig.~\ref{fig:1} (a), is \emph{immediately} and \emph{greatly} reduced (\eg, 65.3$\rightarrow$29.0$\rightarrow$7.2$\rightarrow$4.1), without changing the tracking model, and the GPU/CPU speeds are largely increased as revealed in Fig.~\ref{fig:1} (c). Such an observation is seen in many other Transformer trackers (\eg,~\cite{ye2022joint,lin2022swintrack,cui2022mixformer}), and hence they usually devise a low-resolution speed-oriented variant for efficient visual tracking. 

Albeit faster, simply reducing the input resolution often hurts tracking accuracy much (\eg, 71.0 $\rightarrow$68.7$\rightarrow$62.2$\rightarrow$59.0 in success score), as shown in Fig.~\ref{fig:1} (b). The major reason resulting in such a performance drop is \emph{spatial information loss} including two aspects. On one hand, downsizing the input image to a smaller scale inevitably loses spatial details in pixels, which may degrade feature extraction~\cite{wang2016studying}. On the other hand, when lowering the resolution for the Transformer tracker, \eg,~\cite{ye2022joint}, the tokens of the input become much coarser using the same patch size (\eg,16$\times$16) and decrease in quantity (\eg, 24$^{2}$$\rightarrow$16$^{2}$$\rightarrow$8$^{2}$$\rightarrow$6$^{2}$). Despite largely saving the computation of attention operations, these coarse-grained tokens are \emph{not} capable of capturing fine-grained information interaction within the image that is crucial for discriminative feature learning for enhancing performance in complicated scenarios~\cite{wang2021not}. One solution is to reduce patch size to allow more fine-grained tokens for interaction as in a high-resolution model, while this leads to increased computation. Considering that a trained high-resolution Transformer tracker has already preserved crucial information by learning from fine-grained token interaction in the image, we pose a question: \emph{\textbf{Can we borrow an off-the-shelf high-resolution model to mitigate information loss in low-resolution Transformer tracking to enhance its accuracy while maintaining high efficiency?}}


\textbf{Our Solution.} We show that the answer is \emph{\textbf{positive}} by introducing a simple and novel \emph{dual knowledge distillation} framework, that allows a low-resolution Transformer tracker to learn fine-grained knowledge from its high-resolution counterpart for improving accuracy without sacrificing speed. The crux of our framework lies in two meticulously devised knowledge distillation modules, including \emph{query-key-value knowledge distillation} (QKV-KD) and \emph{discrimination knowledge distillation} (Disc-KD), across resolutions. Specifically, QKV-KD, from the holistic perspective, allows the low-resolution tracker to implicitly inherit more discriminative features and fine-grained interactions in the image from a high-resolution tracker. Compared with common distillation on feature maps (generated by query/key/value) for Transformer, our QKV-KD enables comprehensive knowledge learning from high-resolution feature in multiple views, resulting in better low-resolution feature representation. Besides, another benefit of QKV-KD is indirect knowledge distillation on attention maps, because query and key are directly leveraged for attentional relation computation, better facilitating knowledge learning from the high-resolution Transformer tracker. To further boost the low-resolution tracking, Disc-KD is applied to improve its discriminative ability. Particularly, Disc-KD, from the target-aware view, works to imitate discriminative region generated from the high-resolution model, and enhances the target-background distinguishing capacity of the low-resolution tracker for better performance. 

With dual knowledge distillation by QKV-KD and Disc-KD, the \textbf{Lo}w-\textbf{Re}solution Transformer tracker, coined as \textbf{LoReTrack}, enjoys not only \emph{high efficiency} owing to reduced computation using a smaller input, but also \emph{enhanced accuracy} via distilling discriminative and fine-grained knowledge from the high-resolution tracker. Please notice that, although knowledge distillation, originated from~\cite{hinton2015distilling}, has been adopted for efficient tracking~\cite{shen2021distilled}, this approach boosts performance of a lighter student tracker with a stronger model.  \emph{\textbf{Differently}}, LoReTrack improves the accuracy of the low-resolution tracker using a high-resolution but \emph{not} a stronger model, which avoids designing new student network in~\cite{shen2021distilled} and is thus more compact. Compared with a recent work~\cite{kou2024zoomtrack} on applying non-uniform resizing on smaller input for efficient Transformer tracking, LoReTrack uniformly resizes the resolution and borrows a high-resolution model for knowledge learning, which is simpler yet shows better results.

We implement LoReTrack on top of the popular one-stream Transformer tracking architecture~\cite{ye2022joint}. In order to validate its effectiveness, we conduct extensive experiments on five benchmarks, including LaSOT~\cite{fan2019lasot}, LaSOT$_{\mathrm{ext}}$~\cite{fan2021lasot}, GOT-10k~\cite{huang2019got}, TrackingNet~\cite{muller2018trackingnet}, and UAV123~\cite{mueller2016benchmark}. LoReTrack, with a relatively low resolution of 256$^{2}$, consistently improves the baseline Transformer tracker with the same input resolution, and achieves competitive or even better accuracy compared to the 384$^{2}$ high-resolution tracker, while running \textbf{52\%} faster and saving \textbf{56\%} MACs, as in Fig.~\ref{fig:1}. Moreover, LoReTrack is highly resolution-scale. With a 128$^{2}$ resolution, it run 25 \emph{fps} on a CPU with 64.9\%/46.4\% SUC scores on LaSOT/LaSOT$_{\mathrm{ext}}$, outperforming all other CPU real-time trackers.

In summary, our contributions are three-fold: \textbf{(i)} We propose LoReTrack, an efficient and accurate method that effectively boosts low-resolution Transformer tracking to bridge the gap with its high-resolution counterpart; \textbf{(ii)} We propose two novel knowledge distillation modules, including QKV-KD and Disc-KD, allowing a low-resolution tracker to learn discriminative and fine-grained information from the high-resolution model; and \textbf{(iii)} Extensive experiments exhibit that, LoReTrack achieves consistent improvements over the baseline and meanwhile runs fast, evidencing its efficacy.

\section{Related Work}

\textbf{Transformer Tracking.} Transformer~\cite{dosovitskiy2020image,vaswani2017attention} has recently greatly advanced tracking for its excellent capacity of dependency modeling in images. Early methods (\eg,~\cite{chen2021transformer,wang2021transformer,yan2021learning,gao2022aiatrack}) mainly leverage Transformer to fuse the features from the search region and template for enhancements. To further improve feature representation, fully Transformer tracking (\eg,~\cite{lin2022swintrack,xie2022correlation}) has been introduced. In these approaches, Transformer is employed for both feature extraction and fusion, leading to better performance. Despite this, in these methods the feature extraction backbones for search region and template are learned separately, which neglects the demand for early interaction between search region and template. To address this, the one-stream Transformer tracking framework (\eg,~\cite{cui2022mixformer,chen2022backbone,ye2022joint}) emerges in recent years. It performs feature extraction and interaction for search region and template in a joint way, exhibiting new state-of-the-arts. Owing to its compact design and excellent performance, one-stream architecture has been employed in many subsequent trackers (\eg,~\cite{gao2023generalized,chen2023seqtrack,wu2023dropmae,wei2023autoregressive}).

\textbf{Efficient Tracking.} High efficiency is of importance for the real-world deployment of visual tracking. In recent years, numerous efforts have been made to reduce the computational load of object tracking. The approaches of~\cite{yan2021lighttrack,borsuk2022fear,kang2023exploring} propose to search or directly leverage a more lightweight network for efficient tracking. The work of~\cite{blatter2023efficient} designs a more efficient attention operation to reduce computational complexity of tracking. The method of~\cite{cui2024mixformerv2} compresses the Transformer backbone for accelerating the tracking inference speed. The approach of~\cite{shen2021distilled} proposes to distill a smaller but faster tracking model from a larger one. The method in~\cite{kou2024zoomtrack} leverages a smaller input with non-uniform resizing for decreasing the computational load of Transformer tracking. Our work shares similar spirit with~\cite{shen2021distilled} in using knowledge distillation for high efficiency, but \emph{\textbf{difference}} is that we distill knowledge from a high-resolution model for efficient one-stream Transformer tracking, yet not a larger model as in~\cite{shen2021distilled} for convolutional network-based tracking. In addition, \emph{\textbf{different}} from~\cite{kou2024zoomtrack} that uses a complicated non-uniform resizing method for the smaller input, LoReTrack uniformly resizes the input and learns knowledge from a high-resolution model for improvement, which is simpler yet shows better results.

\textbf{Knowledge Distillation.} Knowledge distillation (KD) is first introduced in~\cite{hinton2015distilling} for learning a more effective student model under the supervision of the teacher model. Later, it has been leveraged for improving model efficiency by distilling knowledge from a larger model to a relatively smaller model, and demonstrated great potential in various tasks such as image recognition~\cite{lin2022knowledge}, object detection~\cite{chen2017learning}, vision-language learning~\cite{wu2023tinyclip}, semantic segmentation~\cite{he2019knowledge}, video recognition~\cite{ma2022rethinking}, etc. Our work is relevant to but significantly different than~\cite{ma2022rethinking}. Specifically, we focus on improving the accuracy of low-resolution Transformer tracking, while~\cite{ma2022rethinking} for convolutional network-based video recognition.

\section{The Proposed Methodology}

\begin{figure}[!t]
    \centering
    \includegraphics[width=0.95\linewidth]{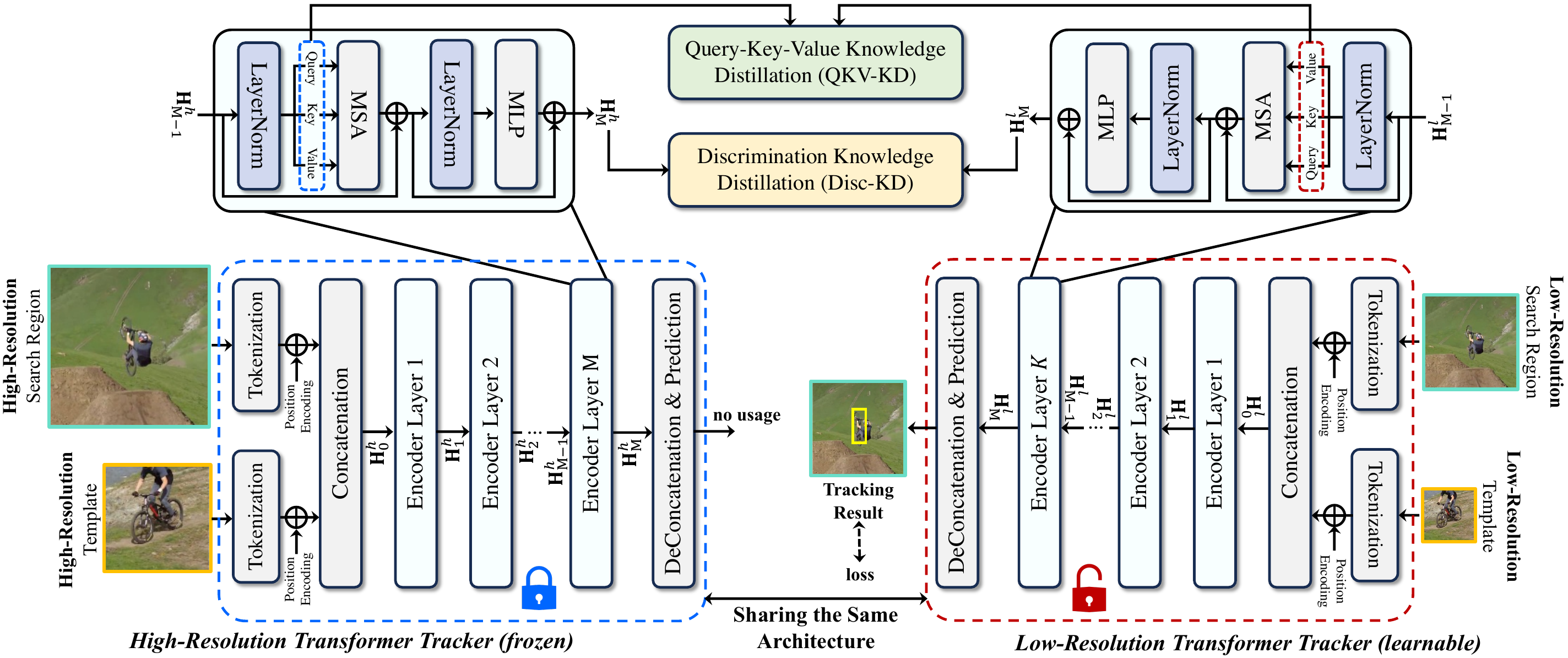}
    \caption{Overview of the proposed method which enhances the accuracy of low-resolution Transformer tracking via dual knowledge distillation from its frozen high-resolution counterpart.}
    \vspace{-10pt}
    \label{fig:framework}
\end{figure}

\textbf{Overview.} In this work, we aim to boost the performance of low-resolution Transformer tracking by distilling knowledge from its frozen high-resolution counterpart while maintaining the efficiency. Specifically, built upon the popular one-stream Transformer tracker~\cite{ye2022joint} ($\S$\ref{preliminary}), we introduce a simple but effective dual knowledge distillation framework ($\S$\ref{dualkd}) that consists of two knowledge distillation modules, including QKV-KD ($\S$\ref{qkv}) and Disc-KD ($\S$\ref{disckd}), for fine-grained and discriminative feature learning. Fig.~\ref{fig:framework} illustrates the overview pipeline of our approach.

\subsection{Base Tracking Architecture}
\label{preliminary}

Our LoReTrack builds upon a popular one-stream Transformer tracking framework OSTrack~\cite{ye2022joint} for its compact architecture and excellent performance. The core idea of OSTrack lies in exploiting the vision Transformer (ViT)~\cite{dosovitskiy2020image} for joint feature extraction and interaction, which enables earlier interaction between the search region and the template for better performance. Specifically, given the template $T$ and search region $S$, we first tokenize them into embeddings $\textbf{E}_{t}=\mathtt{Tokenize}(T)$ and $\textbf{E}_{s}=\mathtt{Tokenize}(S)$, where $\mathtt{Tokenize}(\cdot)$ is the tokenization process consisting of patch partitioning and linear projection. Then, two learnable position embeddings $\textbf{P}_{t}$ and $\textbf{P}_{s}$ are added to $\textbf{E}_{t}$ and $\textbf{E}_{s}$ to generate the final template token embedding $\textbf{T}$ and search region token embedding $\textbf{S}$ as follows,
\begin{equation}
    \textbf{T} = \textbf{E}_{t} + \textbf{P}_{t}  \;\;\;\;\;\; \textbf{S} = \textbf{E}_{s} + \textbf{P}_{s}
\end{equation}
Afterwards, $\textbf{T}$ and $\textbf{S}$ are first concatenated as the final token embedding $\textbf{H}_{0}=\mathtt{Concat}(\mathbf{T}, \mathbf{S})$, where $\mathtt{Concat}(\cdot,\cdot)$ is the concatenation operation. Then,  $\textbf{H}_{0}$ is sent to the Transformer encoder~\cite{dosovitskiy2020image} with $\text{M}$ ($\text{M}=12$) layers. Each layer consists of the multi-head self-attention (MSA) and a multi-layer perceptron (MLP) block for feature interaction and learning. It receives the output from last layer as input (for the first layer, input is $\textbf{H}_{0}$) to perform feature interaction and learning. This process can be expressed as follows,
\begin{equation}
    \begin{split}
        \textbf{Q}_{m} &= \hat{\textbf{H}}_{m} \textbf{W}_{m,q} \;\;\; 
        \textbf{K}_{m} = \hat{\textbf{H}}_{m} \textbf{W}_{m,k} \;\;\;
        \textbf{V}_{m} = \hat{\textbf{H}}_{m} \textbf{W}_{m,v} \;\;\;
        \hat{\textbf{H}}_{m} = \mathtt{LN}(\mathbf{H}_{m-1})
        \\
        \tilde{\textbf{H}}_{m} &= \mathtt{MSA}(\textbf{Q}_{m}, \textbf{K}_{m}, \textbf{V}_{m}) + \mathbf{H}_{m-1}  \\
        \textbf{H}_{m} & = \mathtt{MLP}(\mathtt{LN}(\tilde{\textbf{H}}_{m})) + \tilde{\textbf{H}}_{m} 
    \end{split}
\end{equation}
where $\mathtt{MSA}(\textbf{q},\textbf{k},\textbf{v})$ represents the multi-head attention with query $\textbf{q}$, key $\textbf{k}$, and value $\textbf{v}$, $\texttt{MLP}(\cdot)$ the MLP block, $\mathtt{LN}(\cdot)$ the layer normalization, and $m=1,2,\cdots,\text{M}$. The output $\textbf{H}_{\text{M}}$ from the $\text{M}^{\text{th}}$ layer contains the final features of template and search region. We detach the search region feature $\textbf{F}_{s}$ from $\textbf{H}_{\text{M}}$ by \textcolor{gray}{$\textbf{F}_{t}$}, $\textbf{F}_{s} = \mathtt{DeConcat}(\textbf{H}_{\text{M}})$, where $\mathtt{Deconcat}(\cdot)$ is the deconcatenation operation, and feed it the prediction head that contains classification and regression branches for target localization. 

In this work, both the high-resolution and low-resolution trackers are developed upon OSTrack~\cite{ye2022joint}, and use the same architectures except for different input resolution. For simplicity, we use superscripts ``$h$'' and ``$l$'' to distinguish high- and low-resolution in the following. Please note, candidate elimination in~\cite{ye2022joint} is \emph{not} used as it may result in heavy spatial misalignment in features of different resolutions.

\subsection{Low-Resolution Transformer Tracking with Dual Knowledge Distillation}
\label{dualkd}

To boost the performance of low-resolution Transformer tracking while maintaining its high efficiency, we introduce dual knowledge distillation, which consists of two simple effective modules, including query-key-value knowledge distillation (QKV-KD) and discrimination knowledge distillation (Disc-KD), for learning fine-grained and discriminative features from the high-resolution model. Please note, during knowledge distillation, the high-resolution Transformer tracker is \emph{frozen} after training. 


\subsubsection{Query-Key-Value Knowledge Distillation}
\label{qkv}

\setlength{\columnsep}{10pt}%
\setlength\intextsep{0pt}
\begin{wrapfigure}{r}{0.5\textwidth}
\centering
\includegraphics[width=0.5\textwidth]{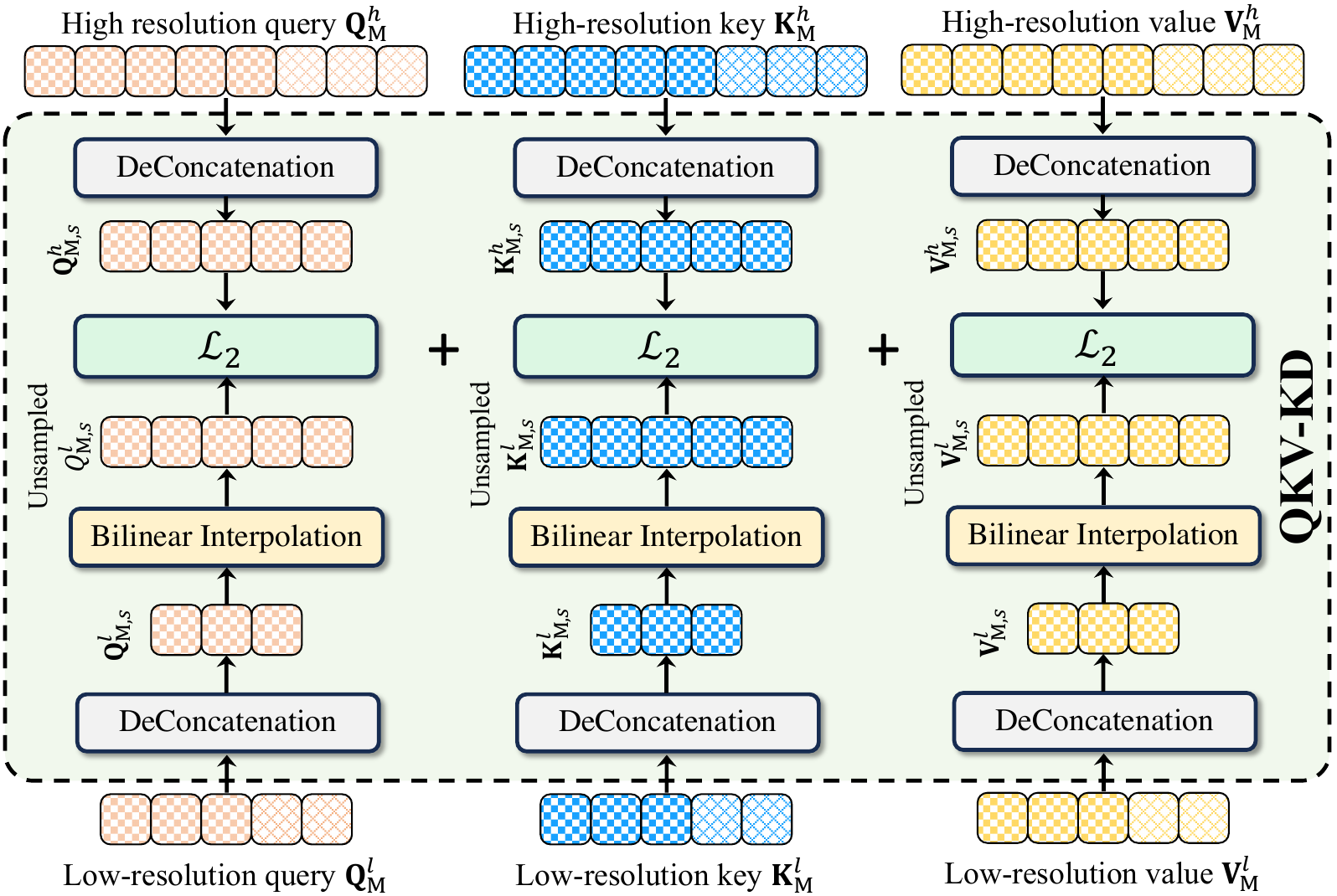}
\caption{Illustration of QKV-KD.}
\label{fig:qkvkd}
\end{wrapfigure}
In low-resolution Transformer tracking,  coarse tokens are difficult for learning discriminative features in the image, which causes accuracy drop. To alleviate this, we devise the \emph{query-key-value knowledge distillation} (QKV-KD), which, from the holistic perspective view, allows the low-resolution Transformer tracker to directly learn discriminative features by fine-grained tokens from the high-resolution model. Different from existing approaches for Transformer that perform knowledge distillation on features (generated by query, key and value)~\cite{lin2022knowledge}, QKV-KD is able to learn more comprehensive knowledge from high-resolution features from \emph{multiple views} (\ie, the query, key, and value views). Besides, another advantage of QKV-KD is that it \emph{indirectly} enables knowledge distillation of the attention maps in Transformer, because query, key, and values are directly leveraged for attention computation, which benefits deeper mimicry for the student low-resolution Transformer tracker~\cite{wang2020minilm}. 

Specifically, we use QKV-KD on the search region in the last $\text{M}^{\text{th}}$ encoder layer  containing the most semantic information, as shown in Fig.~\ref{fig:qkvkd}. More concretely, let us denote $\textbf{H}_{\text{M}-1}^{h}$ and $\textbf{H}_{\text{M}-1}^{l}$ as inputs to the $\text{M}^{\text{th}}$ encoder layer for the high-resolution and low-resolution Transformer trackers. Then, queries, keys, and values, before being sent to MSA, for these two trackers can be generated as follows,
\begin{equation}\small
    \begin{split}
        \textbf{Q}_{\text{M}}^{h} &= \hat{\textbf{H}}^{h}_{\text{M}-1} \textbf{W}_{\text{M},q}^{h} \;\;\;\;\;
    \textbf{K}_{\text{M}}^{h} = \hat{\textbf{H}}^{h}_{\text{M}-1} \textbf{W}_{\text{M},k}^{h} 
    \;\;\;\;\;
    \textbf{V}_{\text{M}}^{h} = \hat{\textbf{H}}^{h}_{\text{M}-1} \textbf{W}_{\text{M},v}^{h} \;\;\;\;\; \hat{\textbf{H}}^{h}_{\text{M}-1} = \mathtt{LN}(\textbf{H}^{h}_{\text{M}-1}) \\
    \textbf{Q}_{\text{M}}^{l} &= \hat{\textbf{H}}^{l}_{\text{M}-1} \textbf{W}_{\text{M},q}^{l} \;\;\;\;\;
    \textbf{K}_{\text{M}}^{l} = \hat{\textbf{H}}^{l}_{\text{M}-1} \textbf{W}_{\text{M},k}^{l} 
    \;\;\;\;\;
    \textbf{V}_{\text{M}}^{l} = \hat{\textbf{H}}^{l}_{\text{M}-1} \textbf{W}_{\text{M},v}^{l} \;\;\;\;\; \hat{\textbf{H}}^{l}_{\text{M}-1} = \mathtt{LN}(\textbf{H}^{l}_{\text{M}-1})
    \end{split}
\end{equation}
where $\textbf{Q}_{\text{M}}^{h}$/$\textbf{K}_{\text{M}}^{h}$/$\textbf{V}_{\text{M}}^{h}$ and $\textbf{Q}_{\text{M}}^{l}$/$\textbf{K}_{\text{M}}^{l}$/$\textbf{V}_{\text{M}}^{l}$ denote queries, keys, and values in the $\text{M}^{\text{th}}$ encoder layers of the low-resolution and high-resolution trackers. Afterwards, we could detach template and search region related queries, keys and values from them via deconcatenation as follows,
\begin{equation}\small
    \begin{split}
        \textcolor{gray}{\textbf{Q}_{\text{M},t}^{h}}, \textbf{Q}_{\text{M},s}^{h} &= \mathtt{DeConcat}(\textbf{Q}_{\text{M}}^{h}) \;\;\;\;\;
        \textcolor{gray}{\textbf{K}_{\text{M},t}^{h}}, \textbf{K}_{\text{M},s}^{h} = \mathtt{DeConcat}(\textbf{V}_{\text{M}}^{h}) \;\;\;\;\;
        \textcolor{gray}{\textbf{V}_{\text{M},t}^{h}}, \textbf{V}_{\text{M},s}^{h} = \mathtt{DeConcat}(\textbf{K}_{\text{M}}^{h})\\
    \textcolor{gray}{\textbf{Q}_{\text{M},t}^{l}}, \textbf{Q}_{\text{M},s}^{l} &= \mathtt{DeConcat}(\textbf{Q}_{\text{M}}^{l}) \;\;\;\;\;
        \textcolor{gray}{\textbf{K}_{\text{M},t}^{l}}, \textbf{K}_{\text{M},s}^{l} = \mathtt{DeConcat}(\textbf{V}_{\text{M}}^{l}) \;\;\;\;\;
        \textcolor{gray}{\textbf{V}_{\text{M},t}^{l}}, \textbf{V}_{\text{M},s}^{l} = \mathtt{DeConcat}(\textbf{K}_{\text{M}}^{l})
    \end{split}
\end{equation}
where $\textbf{Q}_{\text{M},s}^{h}$/$\textbf{K}_{\text{M},s}^{h}$/$\textbf{V}_{\text{M},s}^{h}$ and $\textbf{Q}_{\text{M},s}^{l}$/$\textbf{K}_{\text{M},s}^{l}$/$\textbf{V}_{\text{M},s}^{l}$ are the queries, keys, and values for the search regions in the high-resolution and low-resolution trackers.

With the above, QKV-KD for search region can be mathematically expressed as follows,
\begin{equation}\label{qkvkd}
    \mathcal{L}_{\text{QKV-KD}}=\mathcal{L}_{\text{2}}(\textbf{Q}_{\text{M},s}^{h}, \Phi(\textbf{Q}_{\text{M},s}^{l})) + \mathcal{L}_{\text{2}}(\textbf{K}_{\text{M},s}^{h}, \Phi(\textbf{K}_{\text{M},s}^{l})) + \mathcal{L}_{\text{2}}(\textbf{V}_{\text{M},s}^{h}, \Phi(\textbf{V}_{\text{M},s}^{l}))
\end{equation}
where $\Phi(\cdot)$ denotes the bilinear interpolation operation to align the dimensions, and $\mathcal{L}_2(\cdot,\cdot)$ stands for the mean squared loss. With $\mathcal{L}_{\text{QKV-KD}}$ in Eq. (\ref{qkvkd}), the search region feature of the low-resolution Transformer tracker are able to learn from the discriminative and fine-grained search region feature of the high-resolution Transformer tracker from multiple views, which enhances final performance. Note that, QKV-KD is applied only on the search region. We study using QKV-KD on both template and search region, yet observe no performance gains, as described in experiments later.

\subsubsection{Discrimination Knowledge Distillation}
\label{disckd}

To further foster the discriminative capacity of low-resolution Transformer tracking, we introduce the \emph{discrimination knowledge discrimination} (Disc-KD), which, from the target-aware perspective, focuses on imitating discriminative region generated from the high-resolution model for enhancing the target-background distinguishing ability of the low-resolution tracker.

\setlength{\columnsep}{10pt}%
\setlength\intextsep{0pt}
\begin{wrapfigure}{r}{0.5\textwidth}
\centering
\includegraphics[width=0.5\textwidth]{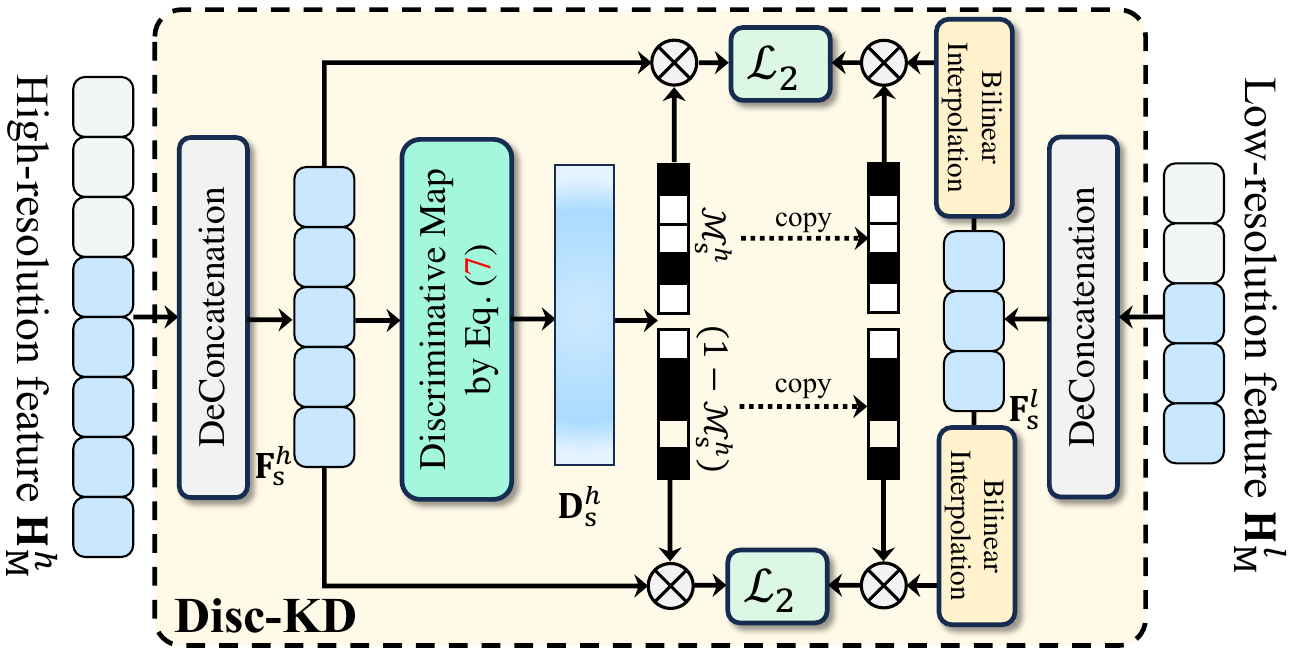}
\caption{Illustration of Disc-KD.}
\vspace{-10pt}
\label{fig:disckd}
\end{wrapfigure}
Considering that the high-level features usually contain more semantic information for discrimination, we apply Disc-KD on the search region feature after the last $\text{M}^{\text{th}}$ encoder layer. Specifically, given the output features $\textbf{H}_{\text{M}}^{h}$ and $\textbf{H}_{\text{M}}^{l}$ after the $\text{M}^{\text{th}}$ encoder layer from the high- and low-resolution trackers, we first detach the search region features from them, as follows,
\begin{equation}
    \begin{split}
        \textcolor{gray}{\textbf{F}_{t}^{h}}, \textbf{F}_{s}^{h} &= \mathtt{DeConcat}(\textbf{H}_{\text{M}}^{h})  \\
        \textcolor{gray}{\textbf{F}_{t}^{l}}, \textbf{F}_{s}^{l} &= \mathtt{DeConcat}(\textbf{H}_{\text{M}}^{l})
    \end{split}
\end{equation}
Inspired by~\cite{zagoruyko2016paying} that the importance of hidden neuron activation can be indicated utilizing its absolute values in feature maps, we generate discriminative maps for the search region using $\textbf{F}_{s}^{h}$ from the high-resolution Transformer model, similar to~\cite{ye2022joint}. In specific, we compute values for each token in $\textbf{F}_{s}^{h}$ along the channel dimension via
\begin{equation}\label{dis}
    \textbf{D}_s^{h}(i) = \frac{1}{C^{h}}\sum_{c=1}^{C^{h}}\big(\textbf{F}_{s}^{h}(i,c)\big)^{2} \;\;\; i=1,2,\cdots,N^{h}
\end{equation}
where $\textbf{D}_s^{h}$ is the discriminative map of $\textbf{F}_{s}^{h}$, $N^{h}$ number of tokens in $\textbf{F}_{s}^{h}$, and $C^{h}$ channel dimension of the token. Note that in Eq. (\ref{dis}), we use square value to compute discriminative map for slightly better performance. 

Instead of directly performing knowledge distillation on $\textbf{F}_s^{h}$ and $\textbf{F}_s^{l}$, we propose to decouple discrimination and non-discrimination regions from features for knowledge distillation, which allows more focus on learning the ability to distinguish target in the low-resolution tracker. More concretely, given $\textbf{D}_s^{h}$, we first extract the discrimination mask $\mathcal{M}_{s}^{h}$ from it with a threshold $\tau$, as follows,
\begin{equation}
	\mathcal{M}_{s}^{h}(i,j)=\left\{
	\begin{aligned}
		1 & , & \textbf{D}_{s}^{h}(i,j) \ge \tau\\
		0 & , & \text{otherwise}
	\end{aligned}
	\right.
\end{equation}
It is worth noticing that, here we employ the discriminative map from the high-resolution tracker to produce the discrimination mask, because the high-resolution feature is more robust than that of low-resolution trackers and the discriminative region is thus more accurate. After this, the Disc-KD can be formulated as follows,
\begin{equation}
    \mathcal{L}_{\text{Disc-KD}}=\alpha_{1}\mathcal{L}_{2}\big(\mathcal{M}_{s}^{h} \otimes \textbf{F}_s^{h}, \mathcal{M}_{s}^{h} \otimes \Phi(\textbf{F}_s^{l})\big) + \alpha_{2}\mathcal{L}_{2}\big((1-\mathcal{M}_{s}^{h}) \otimes \textbf{F}_s^{h}, (1-\mathcal{M}_{s}^{h}) \otimes \Phi(\textbf{F}_s^{l})\big)
\end{equation}
where $\alpha_1$ and $\alpha_2$ are weight parameters to balance the loss. By setting $\alpha_1$ larger than $\alpha_2$, Disc-KD enables paying more attention to distilling knowledge from discriminative target regions.

\subsection{Loss and Inference for Low-Resolution Transformer Tracking}
Similar to OSTrack, after the last $\text{M}^{\text{th}}$ encoder layer, the search region feature $\textbf{F}_{s}^{l}$, which is detached from $\textbf{H}_{\text{M}}^{l}$ by deconcatenation via $\textcolor{gray}{\textbf{F}_{t}^{l}}, \textbf{F}_{s}^{l}=\mathtt{DeConcate}(\textbf{H}_{\text{M}}^{l})$, is reshaped and fed to the prediction head to generate classification scores and regression offsets. During training, we consider the loss terms for classification and regression, as in OSTrack, as well as for knowledge distillation including query-key-value knowledge distillation and discrimination knowledge distillation. Eventually, the total loss $\mathcal{L}_{\text{LoReTrack}}$ for training the low-resolution Transformer tracker can be expressed as follows,
\begin{equation}
    \mathcal{L}_{\text{LoReTrack}} = \mathcal{L}_{\text{Cls}} + \mathcal{L}_{\text{Reg}} + \beta_1\mathcal{L}_{\text{QKV-KD}} + \beta_2\mathcal{L}_{\text{Disc-KD}}
\end{equation}
where $\mathcal{L}_{\text{Cls}}$ and $\mathcal{L}_{\text{Reg}}$ are losses for classification and regression, and $\beta1$ and $\beta_2$ the weight parameters. For details of these two, please kindly refer to OSTrack~\cite{ye2022joint}. Please note that, the high-resolution Transformer tracking model is frozen during the training of LoReTrack.

Once the training of LoReTrack is completed, we perform inference for tracking target object as in the base tracker OSTrack. For further elaboration on this, we kindly refer the readers to~\cite{ye2022joint} 

\renewcommand{\arraystretch}{1.15}
\begin{table}[!t]
\setlength{\tabcolsep}{4.8pt}
  \centering
  \caption{State-of-the-art comparison on five datasets, including LaSOT~\cite{fan2019lasot}, LaSOT$_{\text{ext}}$~\cite{fan2021lasot}, GOT-10k~\cite{huang2019got}, TrackingNet~\cite{muller2018trackingnet}, and UAV123~\cite{mueller2016benchmark}. The speed is measured by \emph{fps}. The best two results are highlighted in \BEST{red} and \SBEST{blue}. `$\P$' indicates the usage of temporal contextual information for improving tracking. Overall, our LoReTrack shows better performance with faster running speed.}
  \scalebox{0.78}{
    \begin{tabular}{rccccccccccccc}
    \midrule[1.3pt]
    \multirow{2}[0]{*}{Tracker} & \multirow{2}[0]{*}{Res.} & \multicolumn{2}{c}{\textbf{LaSOT}} & \multicolumn{2}{c}{\textbf{LaSOT$_\text{ext}$}} & \multicolumn{3}{c}{\textbf{GOT-10k}} & \multicolumn{2}{c}{\textbf{TrackingNet}} & \multicolumn{2}{c}{\textbf{UAV123}} & \multicolumn{1}{c}{\multirow{2}[0]{*}{\makecell{GPU\\ Speed}}} \\
    \cmidrule(l){3-4} \cmidrule(l){5-6} \cmidrule(l){7-9} \cmidrule(l){10-11} \cmidrule(l){12-13}
          &       & SUC   & PRE     & SUC   & PRE     & AO    & SR$_{0.5}$ & SR$_{0.75}$ & SUC   & PRE     & SUC   & PRE     &  \\
    \hline
    LoReTrack-256 (\textbf{ours}) & 256$^{2}$   & \BEST{70.3}  & \BEST{76.2}  & \BEST{51.3}  & \BEST{58.7}  & \BEST{73.5}  & \BEST{84.0}    & \BEST{70.4}  & 82.9  & 81.4  & \BEST{70.6}  &  \BEST{92.1}     & 130 \\
    OSTrack-256~\cite{ye2022joint} & 256$^{2}$   & 68.7  & 74.6  & 46.3    & 51.9     &71.5    & 81.6    & 67.7     & 82.9  & 81.6  & 68.3     & 88.6     & 130 \\
    LoReTrack-192 (\textbf{ours}) & 192$^{2}$   & 68.6  & 73.7  & 50.0    & 56.8  & 71.5  & 82.8  & 66.7  & 80.9  & 78.4  & 69.9  &91.4       & 186 \\
    OSTrack-192~\cite{ye2022joint} & 192$^{2}$   & 66.2  &70.2       & 47.1   &53.0       &67.6       & 78.8   &62.1       &80.7       &77.8       &67.8       &88.5       & 186 \\
    \hdashline
    OSTrack-Zoom~\cite{kou2024zoomtrack} & 256$^{2}$   & \SBEST{70.2}  & \BEST{76.2}  & \SBEST{50.5}  & \SBEST{57.4}  & \BEST{73.5}  & \SBEST{83.6}  & 70.0    & \SBEST{83.2}  & \SBEST{82.2}  & 69.3  & 91.4     & 128 \\
    MixFormerV2~\cite{cui2024mixformerv2}$^{\P}$ & 288$^{2}$   & 69.5  & 75.0    & -     & -     & -     & -     & -     & 82.9  & 81.0    & \SBEST{70.5}  & \SBEST{91.9}  & 165 \\
    OSTrack-CE-256~\cite{ye2022joint} & 256$^{2}$   & 69.1  & 75.2  & 47.4  & 53.3  & 71.0    & 80.4  & 68.2  & 83.1  & 82.0    & 68.3  & 88.8    & 147 \\
    ROMTrack~\cite{cai2023robust}$^{\P}$ & 256$^{2}$   & 69.3  & \SBEST{75.6}  & 48.9  & 55.0  & \SBEST{72.9}  & 82.9  & \SBEST{70.2}  & \BEST{83.6}  & \BEST{82.7}  & -     & -     & 62 \\
    SwinTrack-T~\cite{lin2022swintrack}$^{\P}$  & 224$^{2}$   & 67.2  & 70.8  & 47.6  & 53.9  & 71.3  & 81.9  & 64.5  & 81.1  & 78.4  & 68.8  & -     & 98 \\
    SimTrack-B~\cite{chen2022backbone} & 224$^{2}$   & 69.3  & -     & -     & -     & 68.6  & 78.9  & 62.4  & 82.3  & -     & 69.8  & 89.6  & 40 \\
    TransInMo~\cite{guo2022learning} & 255$^{2}$   & 65.7  & 70.7  & -     & -     & -     & -     & -     & 81.7  & -     & 69.0    & 89.0    & 34 \\
    TransT~\cite{chen2021transformer} & 256$^{2}$   & 64.9  & 69.0    & -     & -     & 72.3  & 82.4  & 68.2  & 81.4  & 80.3  & 69.1  & -     & 50.0 \\
    ToMP-101~\cite{mayer2022transforming}$^{\P}$ & 288$^{2}$   & 68.5  & 73.5  & 45.9  & -     & -     & -     & -     & 81.5  & 78.9  & 66.9  & -     & 20 \\
    \hdashline
    \textcolor[rgb]{ .502,  .502,  .502}{SeqTrack-B~\cite{chen2023seqtrack}}$^{\P}$ & \textcolor[rgb]{ .502,  .502,  .502}{384$^{2}$} & \textcolor[rgb]{ .502,  .502,  .502}{71.5} & \textcolor[rgb]{ .502,  .502,  .502}{77.8} & \textcolor[rgb]{ .502,  .502,  .502}{50.5} & \textcolor[rgb]{ .502,  .502,  .502}{57.5} & \textcolor[rgb]{ .502,  .502,  .502}{74.5} & \textcolor[rgb]{ .502,  .502,  .502}{84.3} & \textcolor[rgb]{ .502,  .502,  .502}{71.4} & \textcolor[rgb]{ .502,  .502,  .502}{83.9} & \textcolor[rgb]{ .502,  .502,  .502}{83.6} & \textcolor[rgb]{ .502,  .502,  .502}{68.6} & \textcolor[rgb]{ .502,  .502,  .502}{-} & \textcolor[rgb]{ .502,  .502,  .502}{15} \\
    \textcolor[rgb]{ .502,  .502,  .502}{OSTrack-CE-384~\cite{ye2022joint}} & \textcolor[rgb]{ .502,  .502,  .502}{384$^{2}$} & \textcolor[rgb]{ .502,  .502,  .502}{71.1} & \textcolor[rgb]{ .502,  .502,  .502}{77.6} & \textcolor[rgb]{ .502,  .502,  .502}{50.5} & \textcolor[rgb]{ .502,  .502,  .502}{57.6} & \textcolor[rgb]{ .502,  .502,  .502}{73.7} & \textcolor[rgb]{ .502,  .502,  .502}{83.2} & \textcolor[rgb]{ .502,  .502,  .502}{70.8} & \textcolor[rgb]{ .502,  .502,  .502}{83.9} & \textcolor[rgb]{ .502,  .502,  .502}{83.2} & \textcolor[rgb]{ .502,  .502,  .502}{70.7} & \textcolor[rgb]{ .502,  .502,  .502}{-} & \textcolor[rgb]{ .502,  .502,  .502}{81} \\
    \textcolor[rgb]{ .502,  .502,  .502}{SwinTrack-B~\cite{lin2022swintrack}}$^{\P}$ & \textcolor[rgb]{ .502,  .502,  .502}{384$^{2}$} & \textcolor[rgb]{ .502,  .502,  .502}{71.3} & \textcolor[rgb]{ .502,  .502,  .502}{76.5} & \textcolor[rgb]{ .502,  .502,  .502}{49.1} & \textcolor[rgb]{ .502,  .502,  .502}{55.6} & \textcolor[rgb]{ .502,  .502,  .502}{72.4} & \textcolor[rgb]{ .502,  .502,  .502}{80.5} & \textcolor[rgb]{ .502,  .502,  .502}{67.8} & \textcolor[rgb]{ .502,  .502,  .502}{84.0 } & \textcolor[rgb]{ .502,  .502,  .502}{82.8} & \textcolor[rgb]{ .502,  .502,  .502}{70.5} & \textcolor[rgb]{ .502,  .502,  .502}{-} & \textcolor[rgb]{ .502,  .502,  .502}{45} \\
    \textcolor[rgb]{ .502,  .502,  .502}{MixFormer-22k~\cite{cui2022mixformer}}$^{\P}$ & \textcolor[rgb]{ .502,  .502,  .502}{320$^{2}$} & \textcolor[rgb]{ .502,  .502,  .502}{69.2} & \textcolor[rgb]{ .502,  .502,  .502}{74.7} & \textcolor[rgb]{ .502,  .502,  .502}{-} & \textcolor[rgb]{ .502,  .502,  .502}{-} & \textcolor[rgb]{ .502,  .502,  .502}{70.7} & \textcolor[rgb]{ .502,  .502,  .502}{80.0} & \textcolor[rgb]{ .502,  .502,  .502}{67.8} & \textcolor[rgb]{ .502,  .502,  .502}{83.1} & \textcolor[rgb]{ .502,  .502,  .502}{81.6} & \textcolor[rgb]{ .502,  .502,  .502}{70.4} & \textcolor[rgb]{ .502,  .502,  .502}{91.8} & \textcolor[rgb]{ .502,  .502,  .502}{25} \\
    \textcolor[rgb]{ .502,  .502,  .502}{AiATrack~\cite{gao2022aiatrack}}$^{\P}$ & \textcolor[rgb]{ .502,  .502,  .502}{320$^{2}$} & \textcolor[rgb]{ .502,  .502,  .502}{69.0} & \textcolor[rgb]{ .502,  .502,  .502}{73.8} & \textcolor[rgb]{ .502,  .502,  .502}{46.2} & \textcolor[rgb]{ .502,  .502,  .502}{-} & \textcolor[rgb]{ .502,  .502,  .502}{69.6} & \textcolor[rgb]{ .502,  .502,  .502}{80.0} & \textcolor[rgb]{ .502,  .502,  .502}{63.2} & \textcolor[rgb]{ .502,  .502,  .502}{82.7} & \textcolor[rgb]{ .502,  .502,  .502}{80.4} & \textcolor[rgb]{ .502,  .502,  .502}{70.6} & \textcolor[rgb]{ .502,  .502,  .502}{-} & \textcolor[rgb]{ .502,  .502,  .502}{38} \\
    \midrule[1.3pt]
    \end{tabular}%
    }
  \label{tab:tab1}%
  \vspace{-10pt}
\end{table}%

\section{Experiments}

\textbf{Implementation details.} The proposed low-resolution Transformer tracker, coined as LoReTrack, is built upon OSTrack~\cite{ye2022joint} and implemented using PyTorch~\cite{paszke2019pytorch}. The parameter settings are kept the same as in the base OSTrack except for the removal of candidate elimination, and please refer to~\cite{ye2022joint} for details. Four datasets, including the training splits of LaSOT~\cite{fan2019lasot}, TrackingNet~\cite{muller2018trackingnet}, COCO~\cite{lin2014microsoft}, and GOT-10k~\cite{huang2019got} (1K forbidden sequences from GOT-10k training set are removed). The threshold $\tau$ is empirically set to 0.2. $\alpha_1$ and $\alpha_2$ are set to 0.6 and 0.4. $\beta_1$ and $\beta_2$ are set to 0.01 and 0.01, respectively. Our model is trained utilizing 4$\times$ NVIDIA A6000 GPUs. The inference is performed on a single NVIDIA A5500 GPU with an Intel i9-11900K CPU.

\textbf{Resolution Setting.} In this work, we use the base OSTrack~\cite{ye2022joint} with 384$^{2}$ input as the high-resolution Transformer. The high-resolution tacker can be directly initialized using off-the-shelf pre-trained model from~\cite{ye2022joint} and used in a frozen way. LoReTrack denotes the base tracker with lower resolutions $\{256^2, 192^2, 128^2, 96^2\}$. Among them, LoReTrack with 256$^{2}$ and 192$^{2}$, \ie, LoReTrack-256 and LoReTrack-192, are designed for low-resolution tracking with GPU, while that with 128$^{2}$ and 96$^{2}$, \ie, LoReTrack-128 and LoReTrack-96, for real-time low-resolution tracking on the CPU. 

\subsection{State-of-the-art Comparison}

We evaluate our LoReTrack-256/192 and compare them with other state-of-the-art trackers on five challenging benchmarks, including LaSOT~\cite{fan2019lasot}, LaSOT$_{\text{ext}}$~\cite{fan2021lasot}, GOT-10k~\cite{huang2019got}, TrackingNet~\cite{muller2018trackingnet}, and UAV123~\cite{mueller2016benchmark}. The success (SUC) and precision (PRE) scores are employed as evaluation metrics on LaSOT, LaSOT$_{\text{ext}}$, TrackingNet, and UAV123, and average overlap (AO) and success rate (SR) for comparison on GOT-10k. \textbf{Please note that}, we mainly compare with trackers with similar resolutions (\eg, smaller than 300$^{2}$), and trackers with higher-resolution (\eg, larger than 300$^{2}$) or using much larger backbones are displayed but excluded in comparison due to different aims. For tracker OSTrack and OSTrack-Zoom, we use the same GPU for speed measurement.

\textbf{LaSOT}~\cite{fan2019lasot} is a large-scale dataset with 280 long-term sequences for testing. As shown in Tab.~\ref{tab:tab1}, our LoReTrack with 256$^{2}$ resolution achieves the best SUC score of 70.3\%, and improves OSTrack-256 with 68.7\% SUC score by 1.6\%. When using the 192$^{2}$ resolution for tracking, LoReTrack-192 improves OSTrack-192 from 66.2\% to 6.6\% with absolute gains of 2.4\%. All these results show the effectiveness of our dual knowledge distillation from the high-resolution tracking model to enhance low-resolution Transformer tracking performance.

\textbf{LaSOT$_{\text{ext}}$}~\cite{fan2021lasot} is an extension of LaSOT by adding 150 new challenging sequences. As demonstrated in Tab.~\ref{tab:tab1}, our LoReTrack achieves the 51.3\% SUC score, which improves the baseline OSTrack-256 by 5.0\%, showing its efficacy. In addition, LoReTrack-256 outperforms the recent OSTrack-Zoom with 50.5\% SUC score by 0.8\%, which displays the advantage of our simple but effective approach. Moreover, LoReTrack outperforms many trackers with higher 384$^{2}$ resolution such as SeqTrack-B, OSTrack-CE, and SwinTrack-B and meanwhile runs faster at 130 \emph{fps}, validating its effectiveness.

\textbf{GOT-10k}~\cite{huang2019got} evaluates the zero-shot performance of trackers. As shown in Tab.~\ref{tab:tab1}, LoReTrack-256 obtains the 73.5\% AO score, which improves OSTrack with the same 256$^2$ resolution with 71.0\% AO score by 2.5\% and outperforms other efficient trackers. Compared to OSTrack-Zoom, LoReTrack achieves the same AO score but better SR$_{0.5}$ and SR$_{0.75}$ scores.

\textbf{TrackingNet}~\cite{muller2018trackingnet} provides 511 videos for evaluation. LoReTrack reveals no improvements in SUC score over the base tracker on TrackingNet, similar to the observation in OSTrack-Zoom. The possible reason is that TrackingNet has a very different test split than other datasets~\cite{kou2024zoomtrack}, and contains very few videos with scenarios that can benefit from looking at fine-grained information.  

\textbf{UAV123}~\cite{mueller2016benchmark} is a challenging drone-based tracking dataset with 123 videos. LoReTrack-256 achieves the SUC score of 70.6\%, which improves OSTrack-256 with 68.3\% by 2.3\%. It surpasses other efficient trackers including the recent OSTrack-Zoom with 69.3\% and MixFormerV2 with 70.5\% SUC scores, and shows competitive performance compared to the high-resolution OSTrack-CE-384.


\renewcommand{\arraystretch}{1.15}
\begin{table}[!t]
\setlength{\tabcolsep}{4.8pt}
  \centering
  \caption{Comparison with CPU real-time tracking algorithms on five datasets, including LaSOT~\cite{fan2019lasot}, LaSOT$_{\text{ext}}$~\cite{fan2021lasot}, GOT-10k~\cite{huang2019got}, TrackingNet~\cite{muller2018trackingnet}, and UAV123~\cite{mueller2016benchmark}. The speed is measured by \emph{fps}. The best two results are highlighted in \BEST{red} and \SBEST{blue}.}
  \scalebox{0.78}{
    \begin{tabular}{rccccccccccccc}
    \midrule[1.3pt]
    \multirow{2}[0]{*}{Tracker} & \multirow{2}[0]{*}{Res.} & \multicolumn{2}{c}{\textbf{LaSOT}} & \multicolumn{2}{c}{\textbf{LaSOT$_\text{ext}$}} & \multicolumn{3}{c}{\textbf{GOT-10k}} & \multicolumn{2}{c}{\textbf{TrackingNet}} & \multicolumn{2}{c}{\textbf{UAV123}} & \multicolumn{1}{c}{\multirow{2}[0]{*}{\makecell{CPU\\ Speed}}} \\
    \cmidrule(l){3-4} \cmidrule(l){5-6} \cmidrule(l){7-9} \cmidrule(l){10-11} \cmidrule(l){12-13}
          &       & SUC   & PRE     & SUC   & PRE     & AO    & SR$_{0.5}$ & SR$_{0.75}$ & SUC   & PRE     & SUC   & PRE     &  \\
    \hline
    LoReTrack-128 (\textbf{ours}) & 128$^2$   & \BEST{64.9}  & \BEST{67.5}  & \BEST{46.4}  & \BEST{50.9}  & \SBEST{64.3}  & \SBEST{76.4}  & 53.6  & \BEST{77.7}  & \BEST{73.6} & \BEST{69.0}    & \BEST{90.8}      & 25 \\
    OSTrack-128~\cite{ye2022joint} & 128$^2$   & \SBEST{62.2}  &\SBEST{63.6}      &43.4   &46.8       &62.9       &75.2       &51.7       &\SBEST{76.6}       &71.5      &66.0       &87.1       & 25 \\
    LoReTrack-96 (\textbf{ours}) & 96$^2$    & 61.0    & 61.1  & \SBEST{45.1}  & \SBEST{49.0}    & 58.9  & 70.9  & 42.2  & 74.0    & 68.3 & \SBEST{67.1}  & \SBEST{88.9}     & 31 \\
    OSTrack-96~\cite{ye2022joint} & 96$^2$    & 59.0    &58.2      &44.0   &47.1      &57.3       &68.7       &40.8       &72.7       &65.9      &64.8       &87.1       & 31 \\
    \hdashline
    HiT-Small~\cite{kang2023exploring} &  128$^2$     &  60.5 & 61.5  &  40.4 & -     &  62.6   & 71.2  & \SBEST{54.4}  & \BEST{77.7} & \SBEST{73.1} & 65.6  & 61.8     & 72 \\
    MixFormerV2-S~\cite{cui2024mixformerv2}$^{\P}$ &  224$^2$     & 60.6  & 60.4  & 43.6  & 46.2 & -     & -     & -     & 75.8  & 70.4 & 65.8  & 86.8  & 30 \\
    FEAR-L~\cite{borsuk2022fear}$^{\P}$ &256$^2$       & 57.9  & 60.9  & -     & -     & 64.5  & -     & -     & -     & - & -     & -     & - \\

    HCAT~\cite{chen2022efficient}  &256$^2$       & 59.3  & 61    &       &       & \BEST{65.1}  & \BEST{76.5}  & \BEST{56.7}  & \SBEST{76.6}  & 72.9 & 62.7  & -     & 45 \\
    
    E.T.Track~\cite{blatter2023efficient} &256$^2$       & 59.1  & -     & -     & -     & -     & -     & -     & 75.0    & 70.6 & 62.3  & -     & 47 \\
    LightTrack~\cite{yan2021lighttrack} & 256$^2$       & 53.8  & 53.7  & -     & -     & 61.1  & 71.0    & -     & 72.5  & 69.5  & -     & -     & 44 \\
    \midrule[1.3pt]
    \end{tabular}%
    }
  \label{tab:realtimeres}%
  \vspace{-10pt}
\end{table}%

\textbf{Comparison with CPU real-time Trackers.} LoReTrack is resolution-scalable, and can be applied to various resolutions for tracking. When further reducing the input resolution to 128$^2$ or 96$^2$, LoReTrack achieves CPU real-time tracking (\ie, tracking speed above 20 \emph{fps} following ~\cite{kristan2020eighth}). Tab.~\ref{tab:realtimeres} shows the results of our CPU real-time LoReTrack-128 and LoReTrack-96 and their comparison with other trackers on five benchmarks including LaSOT~\cite{fan2019lasot}, LaSOT$_{\text{ext}}$~\cite{fan2021lasot}, GOT-10k~\cite{huang2019got}, TrackingNet~\cite{muller2018trackingnet}, and UAV123~\cite{mueller2016benchmark}. LoReTrack-128 achieves superior 64.9\%, 46.4\%, 77.7\%, and 69.0\% SUC scores on LaSOT, LaSOT$_{\text{ext}}$, TrackingNet, and UAV123, and shows competitive performance on GOT-10k.

\subsection{Ablation Study}

In order to further analyze different design choices for LoReTrack, we conduct various ablation studies using LoReTrack-256 on LaSOT~\cite{fan2019lasot} as follows.

\textbf{Impact of QKV-KD and Disc-KD.} The key of LoReTrack lies in two knowledge distillation modules of QKV-KD and Disc-KD. To analyze their roles, we conduct ablations in Tab.~\ref{tb:3(a)}. From Tab.~\ref{tb:3(a)}, we can see that, when removing QKV-KD, the performance drops from 70.3\% to 69.0\% by 1.3\%. When not applying Disc-KD, the SUC score is decreased from 70.3\% to 69.7\% by 0.6\%. These results reveal that both QKV-KD and Disc-KD play a crucial for enhancing the performance of LoReTrack.

\textbf{Impact of the threshold $\tau$.} In Disc-KD, an important parameter is the threshold $\tau$ for discrimination mask generation. We empirically study different values for $\tau$ in Tab.~\ref{tb:3(b)}. As shown in Tab.~\ref{tb:3(b)}, we can see that when setting $\tau$ to 0.20, LoReTrack shows the best performance with 70.3\% SUC score.

\begin{minipage}{\textwidth}
\begin{minipage}[t]{0.33\textwidth}
\makeatletter\def\@captype{table}
        \centering
        \caption{\centering Ablations for QKV-KD and Disc-KD.}
        \begin{tabular}{r c c} 
        \hline
        \multicolumn{2}{r}{{Methods}} &{SUC (\%)}
        \\
        \hline\hline
        \multicolumn{2}{r}{w/o QKV-KD}  &69.0 \\
        \multicolumn{2}{r}{w/o Disc-KD}  &69.7 \\
        \multicolumn{2}{r}{Ours}  &70.3 \\
        \hline
       \end{tabular}
       \label{tb:3(a)}
\end{minipage}
\begin{minipage}[t]{0.33\textwidth}
\makeatletter\def\@captype{table}
\centering
\caption{\centering Ablations for the threshold $\tau$ in Disc-KD.}
        \begin{tabular}{r c c} 
        \hline
        \multicolumn{2}{r}{{Methods}} &{SUC (\%)}
        \\
        \hline\hline
        \multicolumn{2}{r}{$\tau$ = 0.18}  &69.5 \\
        \multicolumn{2}{r}{$\tau$ = 0.20 (ours)}  &70.3 \\
        \multicolumn{2}{r}{$\tau$ = 0.22}  &69.9 \\
        \hline
       \end{tabular}
       \label{tb:3(b)}
\end{minipage}
\begin{minipage}[t]{0.33\textwidth}
\makeatletter\def\@captype{table}
\centering
        \caption{\centering Ablations for the impact of template in query-key-value knowledge distillation.}
        \begin{tabular}{r c c} 
        \hline
        \multicolumn{2}{r}{{Methods}} &{SUC (\%)}
        \\
        \hline\hline
        \multicolumn{2}{r}{w/ template}  &69.7 \\
        \multicolumn{2}{r}{w/o template (ours)}  &70.3 \\
        \hline
       \end{tabular}
       \label{tb:3(d)}
\end{minipage}
\end{minipage}

\begin{minipage}{\textwidth}
\begin{minipage}[t]{0.33\textwidth}
\makeatletter\def\@captype{table}
        \centering
        \caption{\centering Ablations for applying QKV-KD on more layers.}
        \begin{tabular}{r c c} 
        \hline
        \multicolumn{2}{r}{{Layer}} &{SUC (\%)}
        \\
        \hline\hline
        \multicolumn{2}{r}{Layer $\{11, 12\}$}  &69.4 \\
        \multicolumn{2}{r}{Layer $\{12\}$ (ours)}  &70.3 \\
        \hline
       \end{tabular}
       \label{tb:3(e)}
\end{minipage}
\begin{minipage}[t]{0.33\textwidth}
\makeatletter\def\@captype{table}
\centering
        \caption{\centering Ablations for candidate elimination in LoReTrack.}
        \begin{tabular}{r c c} 
        \hline
        \multicolumn{2}{r}{{Methods}} &{SUC (\%)}
        \\
        \hline\hline
        \multicolumn{2}{r}{w/ CE}  &66.8 \\
        \multicolumn{2}{r}{w/o CE (ours)}  &70.3 \\
        \hline
       \end{tabular}
       \label{tb:3(f)}
\end{minipage}
\begin{minipage}[t]{0.33\textwidth}
\makeatletter\def\@captype{table}
        \centering
        \caption{\centering Ablations for different distillation strategies.}
        \begin{tabular}{r c c} 
        \hline
        \multicolumn{2}{r}{{Methods}} &{SUC (\%)}
        \\
        \hline\hline
        \multicolumn{2}{r}{Feature distillation}  &69.4 \\
        \multicolumn{2}{r}{QKV-KD (ours)}  &70.3 \\
        \hline
       \end{tabular}
       \label{tb:3(g)}
\end{minipage}
\end{minipage}

\textbf{Impact of template on knowledge distillation.} To explore the impact of template on QKV-KD, we conduct an experiment in Tab.~\ref{tb:3(d)}. Our result shows that, when performing knowledge distillation on both template and search region in QKV-KD, the accuracy is degraded from 70.3\% to 69.7\% by 0.6\%. The possible reason is that, the search region feature is more crucial for tracking because target is localized in the search region. When performing knowledge distillation for template, knowledge learning from high-resolution model for the search region will be weakened, degrading performance. 

\textbf{Impact of encoder layers for QKV-KD.} QKV-KD aims to distill knowledge from the different views of query, key and value. In the experiment, we explore the impact of encoder layers on QKV-KD in Tab~\ref{tb:3(e)}. As shown in Tab~\ref{tb:3(e)}, we can see that, when applying QKV-KD on more layers, the performance is degraded from 70.3\% to 69.4\% by 0.9\%. The reason is that, when applied on multiple, it may be hard to QKV-KD to effectively learn the high-level knowledge from the high-resolution model. 

\textbf{Impact of candidate elimination.} To avoid heavy spatial misalignment in features, we do not employ candidate elimination (CE). To validate our claim, we conduct an experiment in Tab.~\ref{tb:3(f)}. As shown in Tab.~\ref{tb:3(f)}, we can clearly see that, when applying CE in LoReTrack, the SUC score significantly drops from 70.3\% to 66.8\% by 3.5\%. For this reason, we do not adopt CE in our method. 

\textbf{Impact on different distillation strategies for global features.} To showcase the role of QKV-KD for holistic knowledge learning, we compare it with another version that directly utilizes the feature generated by query, key, and value in Tab~\ref{tb:3(g)}. As shown in Tab~\ref{tb:3(g)}, we observe that distillation on query, key, and value shows better performance, evidencing its effectiveness. 

\begin{wraptable}{r}{0.4\textwidth}
\setlength\intextsep{0pt}
\setlength{\tabcolsep}{2.pt}
	\centering
\renewcommand{\arraystretch}{1.1}
        \caption{Ablations for $\alpha_1$ and $\alpha_2$.}
	\scalebox{1}{
		\begin{tabular}{r c c} 
        \hline
        \multicolumn{2}{r}{{Methods}} &{SUC (\%)}
        \\
        \hline\hline
        \multicolumn{2}{r}{$\alpha_1$ = 0.9, $\alpha_2$ = 0.1}  &70.0 \\
        \multicolumn{2}{r}{$\alpha_1$ = 0.8, $\alpha_2$ = 0.2}  &69.7 \\
        \multicolumn{2}{r}{$\alpha_1$ = 0.7, $\alpha_2$ = 0.3}  &69.7 \\
        \multicolumn{2}{r}{$\alpha_1$ = 0.6, $\alpha_2$ = 0.4 (ours)}  &70.3 \\
        \multicolumn{2}{r}{$\alpha_1$ = 0.5, $\alpha_2$ = 0.5}  &70.0 \\
        \hline
       \end{tabular}}
	\label{tab:ttsasa}
\end{wraptable}
\textbf{Impact on different weight parameters $\alpha_1$ and $\alpha_2$.} The parameters $\alpha_1$ and $\alpha_2$ are used to control the weight of knowledge distillation on discrimination and non-discrimination regions. When $\alpha_1$ is way larger than $\alpha_2$, it indicates the distillation focuses on the discrimination region, while less on the background, which may cause weak ability in suppressing background. We conduct experiments to study different values for $\alpha_1$ and $\alpha_2$. Empirically, when setting $\alpha_1$ and $\alpha_2$ to be 0.6 and 0.4, we show the best performance. 

\section{Conclusion}

In this work, we propose to enhance the low-resolution Transformer tracking performance with dual knowledge distillation. The key lies in two specially designed modules, including query-key-value knowledge distillation and discrimination knowledge distillation. The former aims at learning fine-grained features from the high-resolution model, while the later works to mimic the discrimination regions from its high-resolution counterpart. With dual knowledge distillation, the low-resolution Transformer tracker, named LoReTrack, enjoys both high accuracy by distilling useful knowledge from the high-resolution model and fast speed by reduced resolution. In our extensive experiments, we show that LoReTrack effectively improves the baseline Transformer tracker with the same resolution in most cases. Moreover, LoReTrack is resolution-scalable. When further reducing the resolution, LoReTrack achieves CPU real-time tracking while being accurate, outperforming other methods.

\newpage

{
\small
\bibliographystyle{plain}
\bibliography{main.bib}
}






\end{document}